\documentclass[11pt, authoryear,twocolumn, final]{elsarticle}
\usepackage{amssymb}
\usepackage{pifont}
\usepackage{graphicx}
\usepackage{subfig}
\usepackage{natbib}
\usepackage{amsmath}
\usepackage{algorithm}
\usepackage{algpseudocode}
\usepackage{tikz}
\usepackage{verbatim}

\usetikzlibrary{shapes,arrows,chains}

\usepackage{geometry}
\begin{document}

\begin{frontmatter}

\title{Improving automated multiple sclerosis lesion segmentation with a cascaded 3D convolutional neural network approach}
\author[label1]{Sergi Valverde \corref{corr1}}
\author[label1]{Mariano~Cabezas}
\author[label1]{Eloy~Roura}
\author[label1]{Sandra~Gonz\'alez-Vill\`a}
\author[label2]{Deborah Pareto}
\author[label3]{Joan~C.~Vilanova}
\author[label4]{Llu\'is Rami\'o-Torrent\`a}
\author[label2]{\`Alex~Rovira}
\author[label1]{Arnau~Oliver}
\author[label1]{Xavier~Llad{\'o}}

\address[label1]{Research institute of Computer Vision and Robotics, University of Girona, Spain}
\address[label2]{Magnetic Resonance Unit, Dept of Radiology, Vall d'Hebron University Hospital, Spain}
\address[label3]{Girona Magnetic Resonance Center, Spain}
\address[label4]{Multiple Sclerosis and Neuroimmunology Unit, Dr. Josep Trueta University Hospital, Spain}

\cortext[corr1]{Corresponding author. S. Valverde, Ed. P-IV, Campus Montilivi, University of Girona, 17071 Girona (Spain).
e-mail: svalverde@eia.udg.edu. Phone: +34 972 418878; Fax: +34 972 418976.}

\begin{abstract}

  In this paper, we present a novel automated method for White Matter (WM) lesion segmentation of Multiple Sclerosis (MS) patient images. Our approach is based on a cascade of two 3D patch-wise convolutional neural networks (CNN). The first network is trained to be more sensitive revealing possible candidate lesion voxels while the second network is trained to reduce the number of misclassified voxels coming from the first network. This cascaded CNN architecture tends to learn well from small sets of training data, which can be very interesting in practice, given the difficulty to obtain manual label annotations and the large amount of available unlabeled Magnetic Resonance Imaging (MRI) data. We evaluate the accuracy of the proposed method on the public MS lesion segmentation challenge MICCAI2008 dataset, comparing it with respect to other state-of-the-art MS lesion segmentation tools. Furthermore, the proposed method is also evaluated on two private MS clinical datasets, where the performance of our method is also compared with different recent public available state-of-the-art MS lesion segmentation methods. At the time of writing this paper, our method is the best ranked approach on the MICCAI2008 challenge,  outperforming the rest of 60 participant methods when using all the available input modalities (T1-w, T2-w and FLAIR), while still in the top-rank (3rd position) when using only T1-w and FLAIR modalities. On clinical MS data, our approach exhibits a significant increase in the accuracy segmenting of WM lesions when compared with the rest of evaluated methods, highly correlating ($r\ge0.97$) also with the expected lesion volume. 

\end{abstract}

\end{frontmatter}

\newpage
\section{Introduction}
\label{sec:introduction}

Multiple Sclerosis (MS) is the most common chronic immune-mediated disabling neurological disease affecting the central nervous system \citep{Steinman1996}. MS is characterized by areas of inflammation, demyelination, axonal loss, and the presence of lesions, predominantly in the white matter (WM) tissue \citep{Compston2008}. Nowadays, magnetic resonance imaging (MRI) is extensively used in the diagnosis and monitoring of MS \citep{Polman2011}, due to the sensitivity of structural MRI disseminating WM lesions in time and space \citep{Rovira2015,Filippi2016}. Although expert manual annotations of lesions is feasible in practice, this task is both time-consuming and prone to inter-observer variability, which has been led progressively to the development of a wide number of automated lesion segmentation techniques \citep{Llado2012, Garcia-Lorenzo2013}.

Among the vast literature in the field, recent techniques proposed for MS lesion segmentation include supervised learning methods such as decision random forests \citep{Geremia2011, Jesson2015}, ensemble methods \citep{Cabezas2014}, non-local means \citep{Guizard2015}, k-nearest neighbors \citep{Steenwijk2013,Fartaria2015} and combined inference from patient and healthy populations \citep{Tomas-Fernandez2015}. Several unsupervised methods have been also proposed, based either in probabilistic models \citep{Harmouche2015, Strumia2016} and thresholding methods with post-processing refinement \citep{Schmidt2012, Roura2015}. 

During the last years, a renewed interest in deep neural networks has been observed. Compared to classical machine learning approaches, deep neural networks require lower manual feature engineering, which in conjunction with the increase in the available computational power -mostly in graphical processor units (GPU)-, and the amount of available training data, make these type of techniques very interesting  \citep{LeCun2015}. In particular, convolutional neural networks (CNN) have demonstrated breaking performance in different domains such as computer vision semantic segmentation \citep{Simonyan2014} and natural language understanding \citep{Sutskever2014}.

CNNs have also gained popularity in brain imaging, specially in tissue segmentation \citep{Zhang2015, Moeskops2016} and brain tumor segmentation \citep{Kamnitsas2016, Pereira2016, Havaei201718}. However, only a few number of CNN methods have been introduced so far to segment WM lesions of MS patients. \citet{Brosch2016} have proposed a cross-sectional MS lesion segmentation technique based on deep three-dimensional (3D) convolutional encoder networks with shortcut connections and two interconnected pathways. Furthermore, \citet{Havaei2016} have also introduced another lesion segmentation framework with independent image modality convolution pipelines that reduces the effect of missing modalities of new unseen examples.
In both cases, authors reported a very competitive performance of their respective methods in public and private data such as the MS lesion segmentation challenge MICCAI2008  database\footnote{http://www.ia.unc.edu/MSseg}, which is nowadays considered as a performance benchmark between methods. 

\begin{figure*}[tp]
  \vspace{-1cm}
  \begin{center}
    \hspace{-1cm}
    \subfloat[Cascade based pipeline]{
      \includegraphics[width=0.5\textwidth]{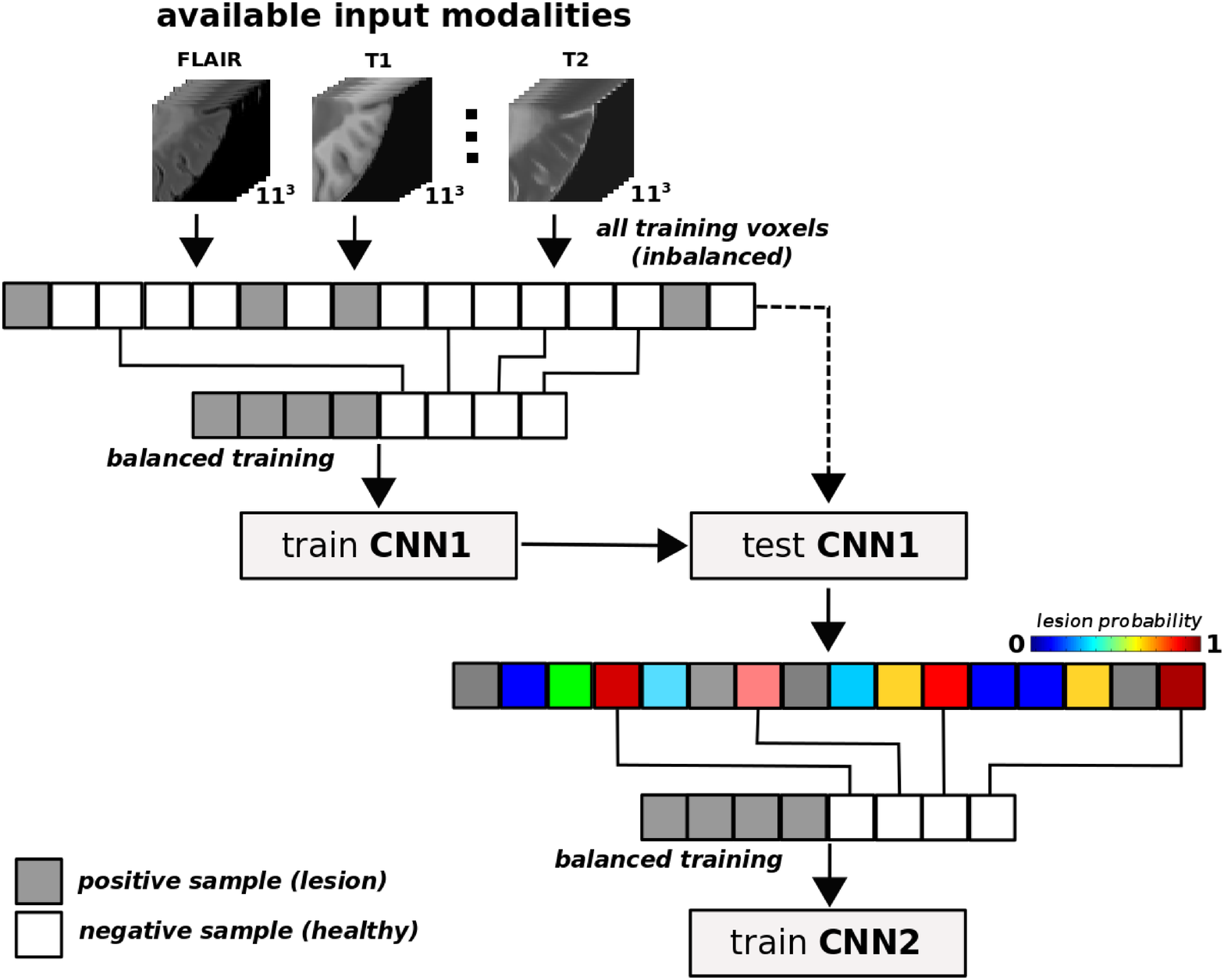}
    }
    \subfloat[Proposed CNN architecture]{
      \hspace{0.5cm}
      \includegraphics[width=0.5\textwidth]{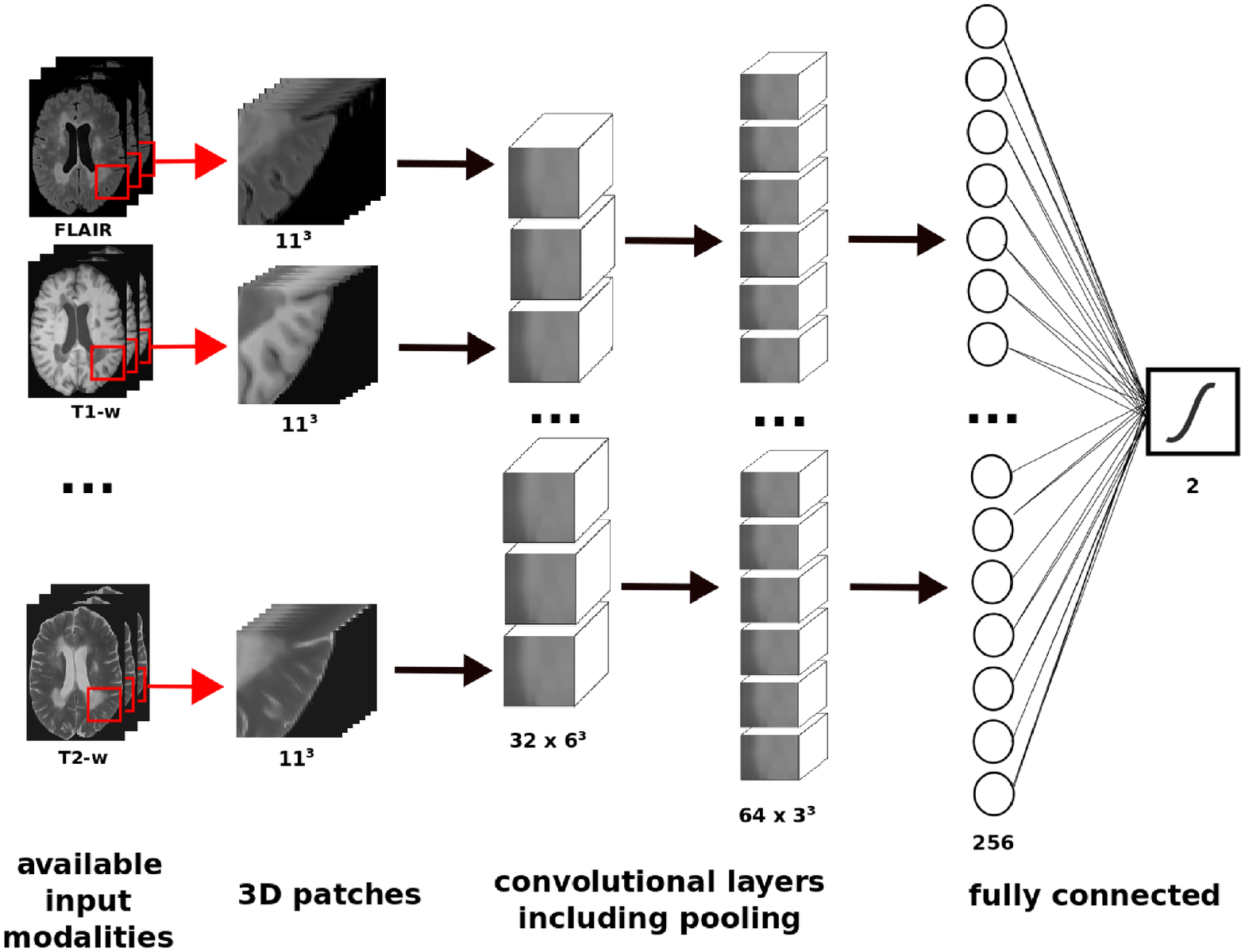}
    }
  \end{center}
  \caption{Proposed pipeline for WM lesion segmentation. (a) Cascade based pipeline, where the output of the first CNN is used to select the input features of the second CNN. (b) The proposed 7-layer CNN model is trained using multi-sequence 3D image patches sampled from a set of training images, where each channel is created from each of the image sequences available in the training database, but not restricted to. See Table 1 for details about each layer parameters.}
    \label{CNN_pipeline}
\end{figure*}

In this paper, we present a new pipeline for automated WM lesion segmentation of MS patient images, which is based on a cascade of two convolutional neural networks. Although similar cascaded approaches have been used with other machine learning techniques in brain MRI \citep{Moeskops2015,Wang2015}, and also in the context of CNNs for coronary calcium segmentation \citep{Wolterink2016a}, to the best of our knowledge this is the first work proposing a cascaded 3D CNN approach for MS lesion segmentation. Within our approach, WM lesion voxels are inferred using 3D neighboring patch features from different input modalities. The proposed architecture builds on an initial prototype that we presented at the recent Multiple Sclerosis Segmentation Challenge (MSSEG2016) \citep{Commowick2016}\footnote{https://portal.fli-iam.irisa.fr/msseg-challenge/workshop-day}. That particular pipeline showed very promising results, outperforming the rest of participant methods in the overall score of the challenge. However, the method presented here has been redesigned based on further experiments to determine optimal patch size, regularization and post-processing of lesion candidates. As in previous studies \citep{Roura2015, Guizard2015, Strumia2016, Brosch2016}, we validate our approach with both public and private MS datasets. First, we evaluate the accuracy of our proposed method with the MICCAI2008  public dataset to compare the performance with respect to state-of-the-art MS lesion segmentation tools. Secondly, we perform an additional evaluation on two private MS clinical datasets, where the performance of our method is also compared with different recent public available state-of-the-art MS lesion segmentation methods. 

\section{Methods}
\label{sec:materials_methods}

\subsection{Input features}
\label{input_features}

Training features are composed by multi-channel 3D patches sampled from a set of training images, where a new channel is created for each input image sequence available in the training database, but not restricted to. 3D feature patches take advantage of the spatial contextual information in MRI. 3D feature patches take advantage of the spatial contextual information in MRI, which may be beneficial in WM lesion segmentation due to the 3D shape of WM lesions \citep{Rovira2015}. The following steps are used to select the input features: 

\begin{enumerate}[i)]

\item Each training image is first subtracted by its mean and divided from its variance, in order to normalize image intensities across different images and speed-up training \citep{LeCun1998}. 

\item For each training image, we compute 3D axial patches of size $p$ centered on the voxel of interest, where $p$ denotes the size in each of the dimensions. 

\item The set of all computed patches $P$ is stacked as $P = [ n \times c \times p \times p \times p]$, where $n$ and $c$ denote the number of training voxels and available input modalities, respectively.
\end{enumerate}

In all experiments, input patch size $p$ is set to $p=11$. This value has been empirically selected after performing different optimization experiments with several patches sizes equal to $p=\{9, 11, 13$,$15\}$. 

\subsection{Cascade based training}

When large amounts of training data are available, several studies have shown that the addition of more CNN layers improves the accuracy in neural networks classification tasks \citep{Simonyan2014,He2015}, usually followed by an additional increase in the number of parameters in comparison to shallower networks \citep{Krizhevsky2012}. However, in MS lesion segmentation, the number of available images with labeled data may be limited by the high number of MRI slices to annotate manually \citep{Llado2012}. More importantly, from the entire number of available voxels, only a very small number of those are lesions ($\sim 1.5\%$ of total brain volume for a patient with 20ml lesion volume), which drastically reduces the number of positive training samples.

These limitations clearly affect the designed architecture, as CNNs tends to suffer from overfitting when they are not trained with enough data. In this aspect, our particular approach has been designed to deal with these two issues. By adequately sampling the training data and splitting the training procedure into two different CNN networks, we design a pipeline with fewer parameters while not compromising the precision and the accuracy of segmenting WM lesions. The next points present in detail each of the necessary steps used by our cascade based training procedure. For a more graphical explanation, see Figure \ref{CNN_pipeline}(a). 

\begin{enumerate}[i)]

\item First, the set of input patches $P = [n \times c \times p \times p \times p]$ is computed from the available training images and input modalities, as described in Section \ref{input_features}. An additional list of labels $L_n$ is also computed using the manual expert lesion annotations, where $L_n=1$ if the current voxel $n$ is a lesion and $L_n=0$ afterwards.  

\item From the entire set $P$, patches where the intensity of the voxel of interest in the FLAIR channel is ($i_n^{FLAIR} < 0.5)$ are removed. WM lesions are placed in either WM or Gray Matter (GM) boundaries, so by simply thresholding voxels on the negative class to signal intensities $i_n^{FLAIR} < 0.5$, we increase the chance of more challenging negatives examples after sampling.

\item In order to deal with data imbalance, we randomly under-sample the negative class in $P$. The training feature set $F_1$ is composed by all positive patches (WM lesion voxels) and the same number of negative patches (healthy voxels) after randomly sampling them. 

\item The network ($CNN_1$) is trained using the balanced training feature set $F_1$. Exact details of the CNN architecture and training are described in Sections \ref{cnn_architecture} and \ref{cnn_training}, respectively.

\item All patches in $P$ are afterwards evaluated using the already trained $CNN_1$, obtaining the probability $Y^1_n$ of each voxel $n$ to belong to the positive class.

\item Based on $Y_n^1$, a new balanced training feature set $F_2$ is created by using again all positive voxels in $P$ and the same number of randomly selected negative examples that have been misclassified in $Y^1_n$, so $Y^1_n > 0.5$ with $L_n = 0$.

\item Finally, the second $CNN_2$ is trained from scratch using the balanced training feature set $F_2$. The output of the $CNN_2$ is the probability $Y_n^2$ for each voxel of being part of a WM lesion. 
  
\end{enumerate}

\subsection{CNN architecture}
\label{cnn_architecture}

Although increasingly deep CNN architectures have been proposed in brain MRI lately \citep{Chen2016a, Cicek2016, Moeskops2016c},  still those tend to be shallower than other computer vision CNN architectures proposed for object recognition or image segmentation of natural images  with up to 150 layers \citep{He2015}, mostly due to factors such as a lower number of training samples, image resolution, and a poorer contrast between classes when compared to natural images. However, compared with the latter, the lower variation in MRI tissues permits the use of smaller networks, which tends to reduce overfitting, specially in small training sets. Here, a 7-layer architecture is proposed for both $CNN_1$ and $CNN_2$ (see Figure \ref{CNN_pipeline}b). Each network is  composed by two stacks of convolution and max-pooling layers with 32 and 64 filters, respectively. Convolutional layers are followed by a fully-connected (FC) layer of size 256 and a soft-max FC layer of size 2 that returns the probability of each voxel to belong to the positive and negative class.  Exact parameters of each of the layers are shown in Table \ref{cnn_parameters}.

\begin{table}[tp]
  \tiny
  \caption{Proposed 7-layer CNN architecture for input image patch size of $11\times 11 \times 11$ with $c$ input modalities as channels. Layer description: 3D convolutional layer (CONV), 3D max-pooling layer (MP) and fully-convolutional layer (FC). Same architecture is proposed for both CNNs.}
  \label{cnn_parameters}
  \begin{center}
    \begin{tabular}{cllcccc}
       \hline
       Layer & Type       & Input size                      & Maps & Size & Stride & Pad\\
       \hline
       0 & \emph{input} & $c \times 11 \times 11 \times 11$  &             &       &      &     \\
       1 & CONV         & $c \times 11 \times 11 \times 11$  & 32          & $3^3$ & $1^3$ & $1^3$ \\
       2 & MP           & $32 \times 5 \times 5 \times 5$ & -           & $2^3$ & $2^3$ & 0\\
       3 & CONV         & $64 \times 5 \times 5 \times 5$ & 64          & $3^3$ & $1^3$ & $1^3$\\
       4 & MP           & $64 \times 2 \times 2 \times 2$ &-            & $2^3$ & $2^3$ & 0 \\
       5 & FC           & 256                             & 256          & 1    & -     & -\\
       6 & Softmax           &  2                              &2             & 1    & -     &  -\\
       \hline
     \end{tabular}
   \end{center}
 \end{table}

\subsection{CNN training}
\label{cnn_training}

To optimize CNN weights, training data is split into training and validation sets. The training set is used to adjust the weights of the neural network, while the validation set is used to measure how well the trained CNN is performing after the epoch, defined as a measure of the number of times all of the training samples are used once to update the architecture's weights. Each CNN is trained individually without parameter sharing. The rectified linear activation function (ReLU) \citep{Nair2010} is applied to all layers. All convolutional layers are initialized using the method proposed by \citet{Glorot2010}. Network parameters are learned using the adaptive learning rate method (ADADELTA) proposed by \citet{Zeiler2012} with batch size of 128 and categorical cross-entropy as loss cost. In order to reduce over-fitting, batch-normalization regularization \citep{Ioffe2015} is used after both convolutional layers, and Dropout \citep{Srivastava2014} with ($p=0.5$) before the first fully-connected layer. Additionally, the CNN model is implemented with early stopping, which permits also to prevent over-fitting by stopping training after a number of epochs without a decrease in the validation error.  Hence, final network parameters are taken from the epoch with the lowest error before stopping. 

\subsection{Data augmentation}

Data augmentation has been shown to be an effective method to increase the accuracy of CNN networks in brain MRI \citep{Pereira2016, Havaei201718, Kamnitsas2016}. Following a similar approach, we perform data augmentation on-the-fly at batch time by multiplying the number of training samples by four following the next transformations: for each mini-batch, all patches are first rotated with 180 degrees in the axial plane. From the original and rotated versions of the patches, new versions are computed by flipping those horizontally. Other rotations than 180 degrees are avoided, in order to roughly maintain the symmetry of the brain and avoid artificial rotations of brain structures. In our empirical evaluations, rotated patches were found to increase the segmentation accuracy of the proposed method in $\sim1.5\%$ when compared to non-rotated patches.
 
\subsection{CNN Testing}

\begin{figure}[t]
  \begin{center}
    \includegraphics[width=0.4\textwidth]{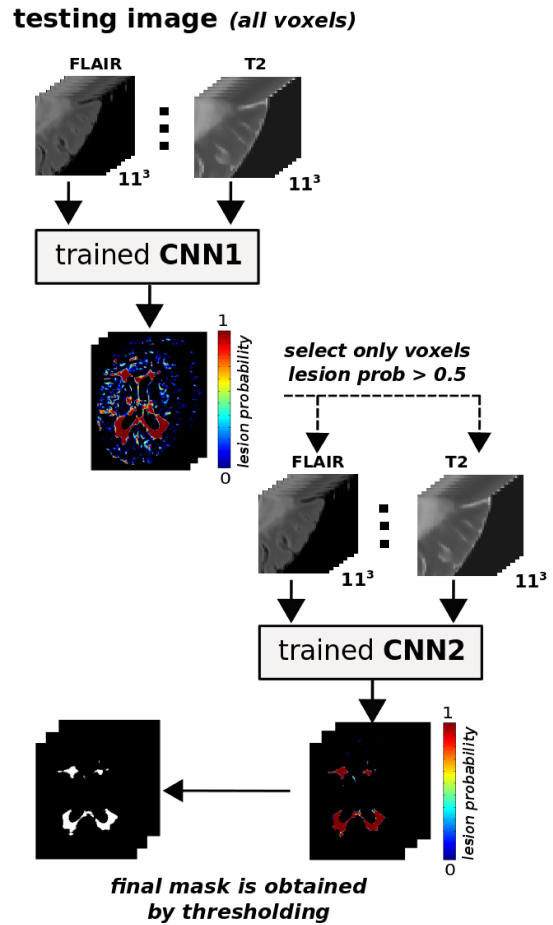}
  \end{center}
    \caption{Proposed CNN testing procedure. New unseen subjects are evaluated using the same cascade architecture of trained networks. Voxels with $\ge 0.5$  probability of being lesion in $CNN_1$ are re-evaluated using the second $CNN_2$. Final segmentation masks are obtained by thresholding the probabilistic mask obtained from the second $CNN_2$. }
    \label{cnn_testing}
\end{figure}

Once the proposed pipeline has been trained, new unseen images are processed using the same CNN cascade architecture (see Figure \ref{cnn_testing}). For each new subject to test, input feature patches for all brain voxels are obtained using the approach proposed in Section \ref{input_features}. All image patches are then evaluated using the first trained CNN. The first network discards all voxels with low probability to be lesion. The rest of the voxels are re-evaluated using the second CNN obtaining the final probabilistic lesion mask. 

Binary output masks are computed by linear thresholding $t_{bin}$ of probabilistic lesion masks. Afterwards, an additional false-positive reduction is performed by discarding binary regions with lesion size below $l_{min}$ parameter. Both parameters $t_{bin}$ and $l_{min}$ are automatically optimized by averaging the best values for each image used for training. Note that using this process, $t_{bin}$ and $l_{min}$ are only computed once after training the network and can be afterwards applied to an arbitrary number of unseen testing images.

\subsection{Implementation}

The proposed method has been implemented in the Python language\footnote{https://www.python.org/}, using Keras\footnote{https://keras.io/} and Theano\footnote{http://deeplearning.net/software/theano/} \citep{Bergstra2011} libraries. All experiments have been run on a GNU/Linux machine box running Ubuntu 14.04, with 32GB RAM memory. CNN training has been carried out on a single Tesla K-40c GPU (NVIDIA corp, United States) with 12GB RAM memory. The proposed method is currently available for downloading at our research website\footnote{https://github.com/NIC-VICOROB/cnn-ms-lesion-segmentation}. 

\section{Experimental Results}

\subsection{MICCAI 2008 MS lesion segmentation}
\label{miccai_dataset}

\subsubsection{Data}
The MICCAI 2008 MS lesion segmentation challenge is composed by 45 scans from research subjects acquired at Children's Hospital Boston (CHB, 3T Siemens) and University of North Carolina (UNC, 3T Siemens Alegra) \citep{Styner2008}. For each subject, T1-w, T2-w and FLAIR image modalities are provided with isotropic resolution of $0.5 \times 0.5 \times 0.5~mm^3$ in all images. Data is distributed in two sets:

\begin{itemize}
\item 20 training cases (10 CHB and 10 UNC) are provided with manual expert annotations of WM lesions from a CHB and UNC expert rater.
\item 25 testing cases (15 CHB and 10 UNC) provided without expert lesion segmentation. 
\end{itemize}

\subsubsection{Evaluation}

The evaluation is done blind for the teams by submitting the segmentation masks of the 25 testing cases to the challenge website\footnote{http://www.ia.unc.edu/MSseg/about.html}. Submitted segmentation masks are compared with manual annotations of both UNC and CHB raters. The evaluation metrics are based on the following scores: 
\begin{itemize}

\item The \% error in lesion volume in terms of the absolute difference in lesion volume ($VD$) between manual annotations masks and output segmentation masks:
  
\begin{equation}
  VD = \frac{|TP_s- TP_{gt}|} {TP_{gt}} \times 100
  \label{eq:vd}
\end{equation}

\noindent where $TP_s$ and $TP_{gt}$ denote the number of segmented voxels in the output and manual annotations masks, respectively. 

\item Sensitivity of the method in terms of the True Positive Rate ($TPR$) between manual lesion annotations and output segmentation masks, expressed in \%:

  \begin{equation}
  TPR = \frac{TP_d}{TP_d+FN_d} \times 100
  \label{eq:tp}
\end{equation}

\noindent where $TP_d$ and $FN_d$ denote the number of correctly and missed lesion region candidates, respectively.

\item False discovery rate of the method in terms of the False Positive Rate ($FPR$) between manual lesion annotations and output segmentation masks, also expressed in \%:

  \begin{equation}
  FPR = \frac{FP_d}{FP_d+TP_d} \times 100
  \label{eq:fp}
\end{equation}

\noindent where $FP_d$ denotes number of lesion region candidates incorrectly classified as lesion.

\end{itemize}

\noindent From these evaluation metrics, a single score is computed to rank each of the participant strategies, being 90 points comparable to human expert performance \citep{Styner2008}.

\begin{table*}[tp]
  \begin{center}
    \scriptsize
    \caption{Segmentation results on the MICCAI 2008 MS lesion segmentation test set. Results are shown for 12 out of 60 participant methods. Mean $VD$, $TPR$ and $FPR$ and final scores are shown split by CHB and UNC raters, as in the original submission website. An overall score of 90 is considered comparable to human performance. \textbf{The best value for each score is depicted in bold.} For a complete ranking of all the participant methods, please refer to the challenge website \texttt{http://www.ia.unc.edu/MSseg/about.html}.}  

      \label{table_miccai}
  \begin{tabular}{r|l|rrr|rrr|r}
    \hline
     & ~ &  \multicolumn{3}{|c|}{\textbf{UNC rater}} & \multicolumn{3}{|c|}{\textbf{CHB rater}} & ~\\
    Rank & Method & VD & TPR & FPR & VD & TPR & FPR & \textbf{Score}\\
    \hline
    - & \textit{Human rater} & -  & -  & -  & -  & -  & - & \textit{90.0} \\ 
    1   & Proposed method (T1-w, FLAIR, T2-w) & 62.5 & 55.5 & 46.8 & \textbf{40.8} & \textbf{68.7} & 46.0 & \textbf{87.12}\\
    2,6  & \citet{Jesson2015}                                   & 46.9 & 43.9 & \textbf{32.3} & 113.4 & 53.5 & \textbf{24.2} & 86.93\\
    3  & Proposed method (T1-w, FLAIR)        & 34.4 & 50.6 & 44.1 & 54.2 & 60.2 & 41.8 & 86.70\\
    4,5  & \citet{Guizard2015}                                  & 46.3 & 47.0 & 43.5 & 51.3 & 52.7 & 42.0 & 86.11\\
    7    & \citet{Tomas-Fernandez2015}                          & \textbf{37.8} & 42.0 & 44.1 & 53.4 & 51.8 & 45.1 & 84.46\\
    8,11 &  \citet{Jerman2016}                                  & 58.1 & \textbf{59.0} & 64.7 & 96.8 & 71.3 & 62.8 & 84.16\\
    10   & \citet{Brosch2016}                                   & 63.5 & 47.1 & 52.7 & 52.0 & 56.0 & 49.8 & 84.07\\
    12   & \citet{Strumia2016}                                  & 56.9 & 37.7 & 34.6 & 113.7 & 42.9 & 30.5 & 83.92\\
    14   & \citet{Roura2015}                                    & 65.2 & 44.9 & 43.2 & 158.9 & 55.4 & 40.5 & 82.34\\
    \hline
  \end{tabular}
\end{center}
\end{table*}

\subsubsection{Experiment details}

Provided FLAIR and T2-w image modalities were already rigidly co-registered to the T1-w space. All images were first skull-stripped using BET \citep{Smith2002} and intensity normalized using N3 \citep{Sled1998} with smoothing distance parameter to 30-50 mm \citep{Boyes2008,Zheng2009}. All training and testing images were then resampled from $512 \times 512 \times 512$ ($0.5\times 0.5\times 0.5mm$) to $256 \times 256 \times 256$ ($1 \times 1 \times 1mm$) to reduce the computational training time. In order to maintain the consistency between image modalities and manual annotation masks, a two-step approach was followed: each of the modalities was first down-sampled to ($256\times 256\times 256$) by local averaging. Then, each image modality in the original space was registered against the same down-sampled image, using the Statistical Parametric Mapping (SPM) toolkit (Estimate and reslice), with normalized mutual information as objective function. The resulting transformation matrix of the FLAIR image was then used to resample also manual annotations to ($256\times 256\times 256$).

For this particular experiment, we trained two different pipelines:

\begin{enumerate}[i)]

\item one trained  using all input image modalities available (T1-w, FLAIR and T2-w).
\item one trained using only T1-w and FLAIR images.
\end{enumerate}

Both pipelines were trained using the 20 available images provided with manual expert annotations, resulting on a balanced training database of 800.000 patches. 25\% of the entire training dataset was set to validation and the rest to train the network. As reported in \citet{Styner2008}, UNC manual annotations were adapted to match closely those from CHB, so in our experiments only CHB annotations were used to train the CNNs. The number of maximum training epochs was set 400 with early-stopping of 50 for each network of the cascade. Test parameters $t_{bin}$
and $l_{min}$ were optimized during training to $t_{bin}=0.8,~l_{min} = 5$ and $t_{bin}=0.7,~l_{min}=5$ in the first and second pipeline, respectively.

\subsubsection{Results}
Table \ref{table_miccai} shows the mean $VD$, $TPR$, $FPR$ and final overall scores of the two proposed pipelines. Our results are compared with other 10 out of 60 other online participant strategies. At the time of writing this paper, our proposed pipeline using all available modalities (T1-w, FLAIR and T2-w) has been ranked in the first position of the challenge, outperforming the rest of participant strategies. Moreover, when only using T1-w and FLAIR images, our pipeline was still very competitive and it was ranked in the top three, only outperformed by the approach proposed by \citet{Jesson2015}. 

In terms of $VD$, the proposed method returned the lowest absolute difference in lesion volume of all participants for either UNC manual annotations (using T1-w and FLAIR) or CHB annotations (using T1-w, FLAIR and T2-w). Additionally, our CNN approach depicted a very high sensitivity detecting WM lesions ($TPR$), being only outperformed by \citet{Jerman2016}. As seen in Table \ref{table_miccai}, other pipelines with high sensitivity such as the same work proposed by \citet{Jerman2016} or the CNN presented in \citet{Brosch2016} tended also to increase the number of false positives. Compared to these methods, our proposed pipeline showed a high $TPR$ while maintaining lower false positive bounds.

\subsection{Clinical MS dataset}
\label{clinical_dataset}

\subsubsection{Data}
The MS cohort is composed by 60 patients with clinically isolated syndrome from the same hospital center (Hospital Vall d'Hebron, Barcelona, Spain). In all patients, the initial clinical presentation was clearly suggestive of MS:

Patients were scanned on a 3T Siemens with a 12-channel phased-array head coil (Trio Tim, Siemens, Germany), with acquired input modalities: 1) transverse PD and T2-w fast spin-echo (TR=2500 ms, TE=16-91 ms, voxel size=0.78$\times$0.78$\times$3mm$^3$); 2) transverse fast T2-FLAIR (TR=9000 ms, TE=93 ms, TI=2500 ms, flip angle=120$^{\circ}$, voxel size=0.49$\times$0.49$\times$3mm$^3$); and 3) sagittal 3D T1 magnetization prepared rapid gradient-echo (MPRAGE) (TR=2300 ms, TE=2 ms, TI=900 ms, flip angle=9$^{\circ}$; voxel size=1$\times$1$\times$1.2mm$^3$). In 25 out of the 60 subjects, the PD image modality was not available, so data was separated in two datasets MS1 and MS2 as follows:

\begin{itemize}

\item MS1 dataset: 35 subjects containing T1-w, FLAIR and PD modalities. Within this dataset, WM lesion masks were semi-automatically delineated from PD using JIM software (\texttt{Xinapse Systems, http://www.xinapse.com/home.php}) by an expert radiologist of the same hospital center. Mean lesion volume was $2.8\pm2.5$ ml (range 0.1-10.7 ml). 

\item MS2 dataset: 25 patients containing only T1-w and FLAIR input modalities. WM lesion masks were also semi-automatically delineated from FLAIR using JIM software by the same expert radiologist. Mean lesion volume was $4.1\pm4.7$ ml (range 0.2-18.3 ml).

\end{itemize}

\subsubsection{Evaluation}

Evaluation scores proposed in section \ref{miccai_dataset} are complemented with the following metrics:  

\begin{itemize}

\item The overall \% segmentation accuracy in terms of the Dice Similarity Coefficient ($DSC$) between manual lesion annotations and output segmentation masks:
  
\begin{equation}
  DSC = \frac{2\times{TP_s}}{FN_s+FP_s+2\times{TP_s}} \times 100
  \label{eq:dsc}
\end{equation}

\noindent where $TP_s$ and $FP_s$ denote the number of voxels correctly and incorrectly classified as lesion, respectively, and $FN$ denotes the number of voxels incorrectly classified as non-lesion. 

\item Precision rate of the method expressed in terms of the Positive Predictive Value rate ($PPV$) between manual lesion annotations and output segmentation masks, also expressed in \%:

  \begin{equation}
  PPV = \frac{TP_s}{TP_s+FP_s} \times 100
  \label{eq:fp}
\end{equation}

\noindent where $TP_s$ and $FP_s$ denote the number of correctly and incorrectly lesion region candidates, respectively.

\end{itemize}

\subsubsection{Experiment details}

For each dataset, T1-w and FLAIR images were first skull-stripped using BET \citep{Smith2002} and intensity normalized using N3 \citep{Sled1998} with smoothing distance parameter to 30-50mm. Then, T1-w and FLAIR images were co-registered to the T1-w space using also the SPM toolbox, with normalized mutual information as objective function and tri-linear interpolation with no warping. 

In order to investigate the effect of the training procedure on the accuracy of the method, two different pipelines were considered:

\begin{enumerate}[i)]
\item one trained with leave-one-out cross-validation: for each patient, an independent pipeline was trained from the rest of the patient images of the MS1 and MS2 datasets. Testing parameters $T_{bin},l_{min}$ were computed at global level from all input images used for training. Final values for test parameters were set to $t_{bin}=0.8,~l_{min} =20$. 

\item one trained with independent cross training-testing databases: all images in MS1 were used to train a single pipeline, which was used to evaluate all the images in MS2. Afterwards, the process was inverted and MS2 images were used to train a single pipeline, which was used to evaluate MS1 images. Optimized test parameters $t_{bin}$ and $l_{min}$ were set equal to the previous pipeline. 

\end{enumerate}

\noindent In both experiments, each of the cascaded networks was trained using FLAIR and T1-w image modalities for a maximum number of 400 epochs with early-stopping of 50. For each subject, a training set of approximately  200.000 patches was generated, where 25\% was set to validation and the rest to train the network. The same modalities were used for testing.

\subsubsection{Comparison with other available methods}

The accuracy of the proposed method is compared with two available state-of-the-art MS lesion segmentation approaches such as LST \citep{Schmidt2012} and SLS \citep{Roura2015}. 
Parameters for LST and SLS were optimized by grid-search on the MS1 database. As in \citet{Roura2015}, given that both MS1 and MS2 images were acquired using the same scanner, same parameters were used for MS2. In LST, initial threshold was set to $\kappa=0.15$ and lesion belief map was set to $lgm=gm$. In the case of SLS, smoothing distribution parameter was set to $\alpha=3$, the percentage of GM/WM tissue to $\lambda_{ts}=0.6$, and percentage of neighboring WM to $\lambda_{nb}=0.6$ in the first iteration. In the second iteration, parameters were set to $\alpha=1.7$, $\lambda_{ts}=0.75$ and $\lambda_{nb}=0.7$, respectively. 

\subsubsection{Results}

Table \ref{table_clinical} shows the mean $DSC$, $VD$, $TPR$, $FPR$ and $PPV$ scores for all the evaluated pipelines. As seen in the Table \ref{table_clinical}, our proposed approach clearly outperformed the rest of available MS pipelines by a large margin in terms of detection ($TPR, FPR$), precision ($PPV$) and segmentation ($VD, DSC$). Furthermore, our method yielded a similar performance in both datasets and independently of the training procedure employed, highlighting the capability of the network architecture to detect new lesions on unseen images. 

\begin{table}[h]
  \begin{center}
    \tiny
    \caption{Mean segmentation results for each of the evaluated methods on the two MS clinical datasets. Results are shown for SLS, LST and our proposed approach trained with either leave-one-out experiments (LOU) or different training-testing dataset. Mean $DSC$, $VD$, $TPF$, $FPF$  and $PPV$ are shown for each method and database. The best value for each metric is shown in bold.}
    \label{table_clinical}

    \begin{center}
      \subfloat[MS1 dataset (n=35)]{
        \begin{tabular}{lrrrrr}
          \hline
          \textbf{Method} & \textbf{DSC} & \textbf{VD} & \textbf{TPR} & \textbf{FPR} & \textbf{PPV}\\
          \hline
          SLS \citep{Roura2015} & 33.4 & 81.0 & 49.5 & 38.2   & 61.7 \\
          LST \citep{Schmidt2012} & 32.2 & 49.7 & 44.4 & 41.2 &  58.8 \\
          \hline
          Proposed method (LOU) & \textbf{53.5}& \textbf{30.8} & 77.0 & \textbf{30.5} & \textbf{70.3}\\
          Proposed method (train MS2) & 50.8 & 38.8 & \textbf{79.1} & 35.6 & 65.3\\
          \hline
        \end{tabular}}
       \end{center}

    \begin{center}
      \subfloat[MS2 dataset (n=25)]{
          \begin{tabular}{lrrrrr}
            \hline
            \textbf{Method} & \textbf{DSC} & \textbf{VD} & \textbf{TPR} & \textbf{FPR} & \textbf{PPV}\\
            \hline
            SLS \citep{Roura2015}     & 29.7 & 65.1 & 35.7 & 46.7 & 53.2 \\
            LST \citep{Schmidt2012}   & 27.3 & 59.4 & 58.9 & 40.2 & 59.7 \\
            \hline
            Proposed method (LOU) & \textbf{56.0 }& 27.5 & \textbf{68.2} & 33.6 & 66.1\\
            Proposed method  (train MS1) & 51.9 & \textbf{26.1} & 63.7 & \textbf{27.2} & \textbf{73.0}\\
            \hline
          \end{tabular}}
      \end{center}
    \end{center}
\end{table}

\begin{figure}[t]
  \begin{center}
    \includegraphics[width=0.5\textwidth]{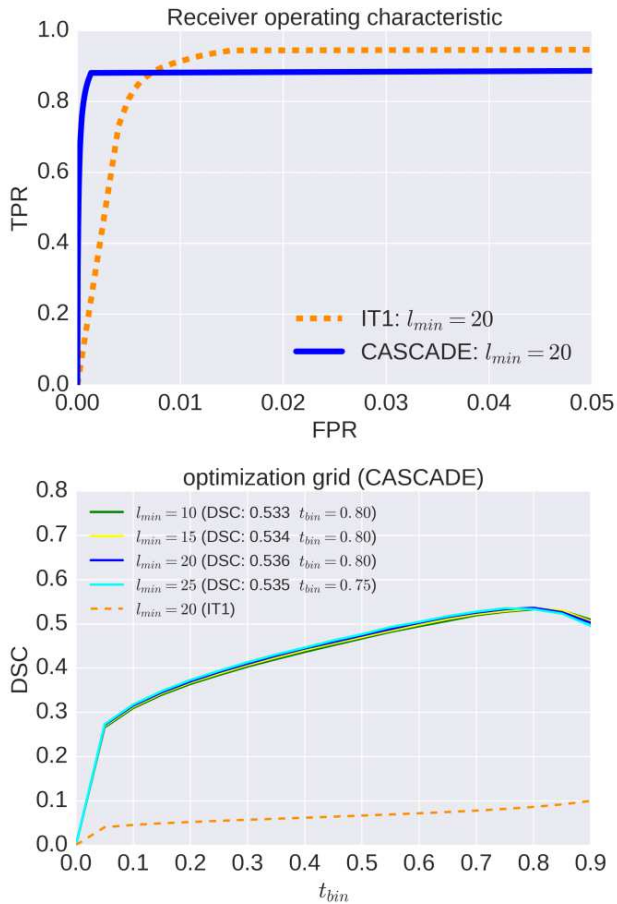}
  \end{center}
    \caption{Receiver Operation Characteristic (ROC) curves and parameter optimization plots on the MS1 database. First row: ROC curves with fixed best minimum lesion size ($l_{min} = 20$) for both the proposed cascaded approach (solid blue) and the same architecture using only the network without cascading (dotted orange). Second row: parameter optimization for the cascaded CNN ($t_{bin}$ and $l_{min}$) against DSC coefficient. Additionally, the best configuration for the CNN approach without cascading is also shown for comparison (dotted orange).}
    \label{fig_roc_curve}
  \end{figure}

Figure \ref{fig_roc_curve} shows the response operating characteristic (ROC) curves and parameter optimization plots for the proposed CNN approach on the MS1 database. Compared to the same CNN architecture without cascading, the proposed approach yielded a higher sensitivity, lower false negative rates and significantly higher DSC overlaps with respect to manual annotations, due to the addition of the second network, which drastically reduced the number of misclassified voxels.

\begin{figure*}[tp]
  \vspace{-1cm}
  \begin{center}
    \includegraphics[width=1\textwidth]{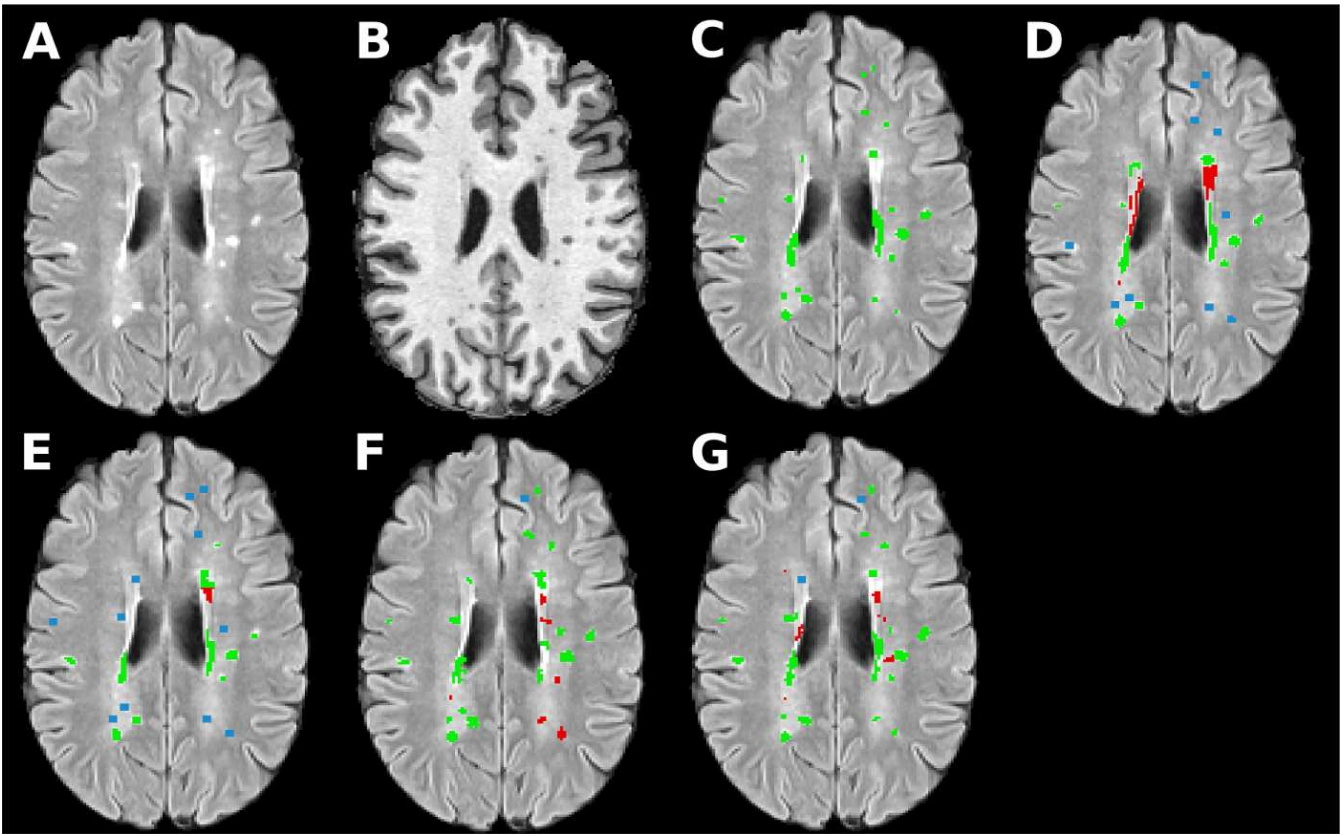}
  \end{center}
    \caption{Output segmentation masks for each of the evaluated methods on the first subject of the MS1 clinical dataset. (A) FLAIR image. (B) T1-w image. (C) Manual WM lesion annotation. Output segmentation masks for SLS (D), LST (E) and our approach when trained using either leave-one-out (F) or the MS2 dataset (E). On all images, true positives are denoted in green, false positives in red and false negatives with a blue square.}
    \label{fig_clinical}
  \end{figure*}

Figure \ref{fig_clinical} depicts a qualitative evaluation of each of the evaluated methods for a single subject of the MS1 dataset. Compared to the SLS and LST pipelines, both proposed CNN pipelines present a significant increase in the number of detected lesions (depicted in green). SLS and LST methods are designed to reduce the $FPR$, so as a counterpart they exhibit a higher number of missed lesions (depicted as blue squares). In contrast, the high sensitivity of the proposed method detecting WM lesions was not compromised by a remarkably increase in the false positives (depicted in red), still lower than the rest of pipelines (see Table \ref{table_clinical}).

  \begin{figure*}[tp]
  \begin{center} 
    \subfloat[MS1 dataset (n=25)]{
      \hspace{-0.5cm}
      \includegraphics[width=1\textwidth]{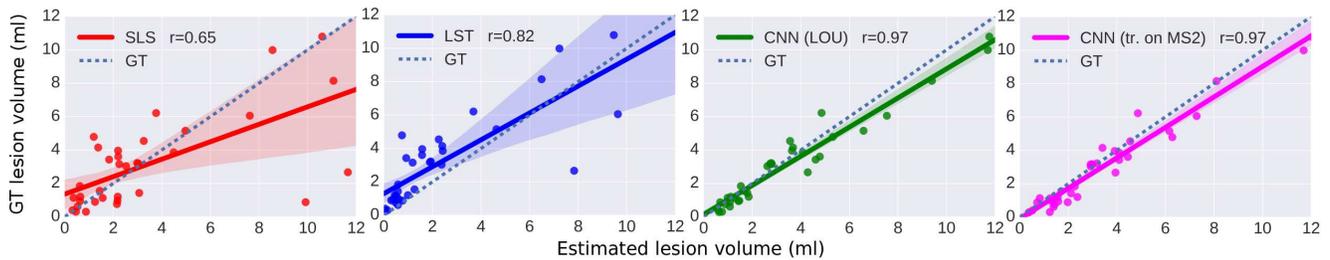}
    }\\
    \subfloat[MS2 dataset (n=35)]{
      \hspace{-0.7cm}
      \includegraphics[width=1\textwidth]{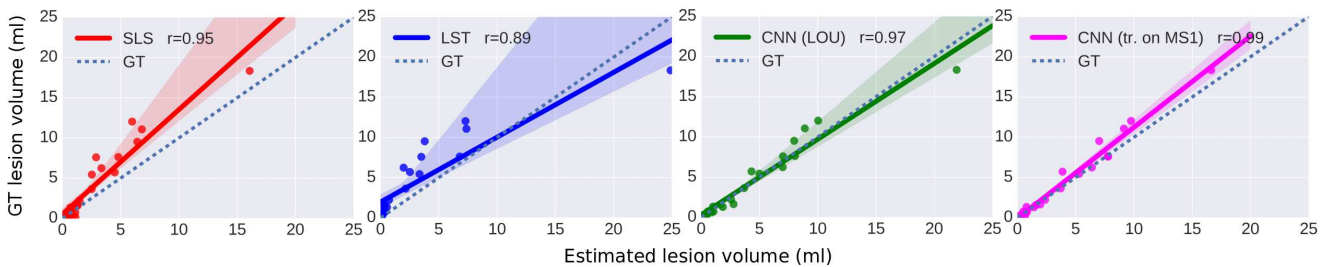}
    }
  \end{center}
    \caption{Correlation between estimated lesion volume and manual WM lesion annotations for each of the evaluated methods on MS clinical databases MS1 (a) and MS2 (b). The linear regression model relating estimated and manual segmentation are plotted (solid lines) along with confidence intervals at 95\%. Estimated models and confidence intervals are compared with respect to expect lesion size (dashed line). For each method and dataset, the Pearson's linear correlations between manual and estimated masks are also shown (p $\le$0.001). }

    \label{corr_lesion_volume}
\end{figure*}

We also compared the correlation between expected and estimated lesion volume for each of the evaluated pipelines and datasets. Figure \ref{corr_lesion_volume} shows the linear regression models fitted based on the volume estimations of each method (solid lines). For each regression model, 95\% confidence interval areas and expected lesion volume (dash lines) were also shown for comparison. In general, distances between expected and computed lesion volumes were lower in CNN architectures when compared to the rest of pipelines, as shown by the Pearson linear correlation coefficients obtained ($r\geq0.97$ in all cases). 
In addition, confidence intervals for our proposed methods were distinctively lower in both datasets, specially in images with higher lesion load, where more variability was introduced.

\label{sec:results}
\section{Discussion}
\label{sec:discussion}

In this paper, we have presented a novel automated WM lesion segmentation method in brain MRI with application to MS patient images. The proposed patch-wise pipeline relies on a cascade of two identical convolutional neural networks, where the first network is trained to be more sensitive revealing possible candidate lesion voxels while the second network is trained to reduce the number of misclassified voxels coming from the first network. Although CNN cascade-based features have been used in coronary calcium segmentation \citep{Wolterink2016a}, still this approach had been not applied in the context of MS lesion segmentation. In our opinion, the proposed cascaded architecture of two CNN is an interesting contribution of the present study. Our experiments have shown that the proposed method outperforms the rest of participant methods in the MICCAI2008 challenge, which is considered nowadays a benchmark for new proposed strategies. Moreover, additional experiments with two clinical MS datasets also confirms our claims about the accuracy of our proposed method. Compared to other state-of-the-art available methods, our approach exhibits a significant increase in the detection and segmentation of WM lesions, highly correlating ($r\ge0.97$) with the expected lesion volume on clinical MS datasets. Furthermore, the performance of the pipeline is consistent when trained with different image datasets, acquisition protocols and image modalities, which highlights its robustness in different training conditions.

Automated WM lesion segmentation pipelines are usually designed as a trade-off between the sensitivity detecting lesions and the capability to avoid false positives, which may affect expected lesion segmentation. In all the experiments handled in this paper, the proposed approach yielded a high sensitivity detecting WM lesions while maintaining the number of false positives reasonably low. Related to that, differences in $VD$ with respect to manual lesion annotations were in average the lowest when using our cascade architecture, specially in images with high lesion size, as seen in the correlation plots between expected and estimated lesion volume of MS patients. In our opinion, the obtained results show the capability of the cascade architecture to reduce false positives without compromising lesion segmentation, which is relevant, as it can have a direct benefit in several tasks such as automated delineation of focal lesions for MS diagnosing \citep{Polman2011}, measuring the impact of a given treatment on WM lesions \citep{Calabrese2012} or reducing the effects of WM lesions in brain volumetry by refilling WM lesions before tissue segmentation \citep{Valverde2014, Valverde2015}. 

In terms of the network architecture, our pipeline has been designed to handle with the lack of large amounts of training data and most importantly, with the imbalanced nature of MRI images of MS patients, where lesion voxels are highly underrepresented. The proposed cascade of two identical CNNs can be also thought as a two-step optimization network, where a more general specific network learns to find candidate lesions while a second more specific network learns to reduce the number of false positives. At testing time, this means that voxels with a low probability to belong to WM lesion are easily discarded by the first network, while challenging voxels flow from the first to the second network, which has been explicitly trained to discriminate between lesion voxels and these challenging non-lesion voxels. Each network is trained independently in different portions of the training data, so the reduced size of each of these networks makes them less prone to overfitting. This makes this particular architecture suitable for small datasets without massive amounts of training samples, as there is a relative low number of parameters to optimize at each of the two steps.

In our proposed design, the sampling procedure followed to deal with data imbalance is equally important. In the first network, the negative class has been under-sampled to the same number of existing lesion voxels, increasing the sensitivity of the network detecting WM lesions at testing time. However, by under-sampling the negative voxel distribution, the probability to misclassify healthy voxels as lesion also increases due to a poorer representation of the negative class. In order to reduce the number of misclassified voxels, the network may be redesigned to incorporate more layers with the aim to learn more abstract features. However, a deeper network increases the number of parameters, and more training data is needed in order to avoid over-fitting. In contrast, within the designed proposal, false positives are reduced by re-training an identical CNN architecture with the most challenging samples derived from the first network.

In contrast to other CNN approaches used in brain MRI \citep{Moeskops2016,Brosch2016}, the training stage is here split into two independent 7-layer CNNs that are exposed to different portions of equal data, so the number of parameters to be optimized at each step is remarkably small ($< 190000$ with input patches of size $11^3$) when compared to other methods. This can be specially interesting when training is performed with a cohort of MS patients with low lesion load, as seen in our experiment containing  only around 200.000 training samples after balancing equally lesion and non-lesion voxels. In this aspect, our results suggest that the proposed pipeline is a valid alternative to train CNN networks without the need of large amounts of training data. In our opinion, this is one of the major contributions of the present study, given the difficulty to obtain labeled MRI data, in comparison to the huge number of available unlabeled data. This suggests that a similar approach may be useful in other similar detection problems like automated segmentation of WM hyperintensities \citep{Griffanti2016},  longitudinal MS lesion segmentation \citep{Cabezas2016}, Lupus lesion detection \citep{Roura2015b}, traumatic brain injury \citep{Lee2005} and brain tumor segmentation \citep{Menze2015}.

The proposed method presents also some limitations. When compared to SLS and LST, although our CNN approach clearly yields  a better accuracy on the MS dataset, this has to be trained and tested for each of the datasets evaluated, which is time consuming and may require more expertise. Related to that, the abstract representations learned by the classifier are most probably provided by the FLAIR and T2-w image sequences, as these sequences are more sensitive revealing the majority of MS lesions when compared with T1-w, for instance. However, the geometry and image contrast of FLAIR / T2-w tend to vary considerably within acquisition protocols. Although changes in intensity scale can be corrected by the internal regularization of the CNN, differences between image domains still may exist, being the feature representations learned by the classifier highly dependent of the dataset used, which suggests us to use it only on the same image domain.

\section{Conclusions}

Automated WM lesion segmentation is still an open field, as shown by the constant number of proposed methods during the last years. In this paper, we have presented a novel automated WM lesion segmentation method with application to MS patient images that relies on a cascade of two convolutional neural networks. Experimental results presented in this paper have shown the consistent performance of the proposed method on both public and private MS data, outperforming the rest of participant methods in the MICCAI2008 challenge, which is considered nowadays as a benchmark for new proposed strategies. Compared to other available methods, the performance of our proposed approach shows a significant increase in the sensitivity while maintaining a reasonable low number of false positives. As a result, the method exhibits a lower deviation in the expected lesion volume in manual lesion annotations with different input image modalities and image datasets. In addition, the proposed cascaded CNN architecture tends to learn well from small sets of data, which can be very interesting in practice, given the difficulty to obtain manual label annotations and the amount number of available unlabeled MRI data. 

The obtained results are encouraging, yielding our CNN architecture closer to human expert inter-rater variability. However, still more research is needed to accomplish this task. Meanwhile, we strongly believe that the proposed method can be a valid alternative for automated WM lesion segmentation.

\section*{Acknowledgements}
This work has been partially supported by "La Fundaci\'{o} la Marat\'{o} de TV3", by Retos de Investigaci\'{o}n TIN2014-55710-R, and by the MPC UdG 2016/022 grant. The authors  gratefully acknowledge the support of the NVIDIA Corporation with their donation of the Tesla K40 GPU used in this research.

\label{sec:conclusion}
\bibliographystyle{apalike}
\bibliography{bibtex}

\end{document}